\documentclass[conference]{IEEEtran}
\IEEEoverridecommandlockouts
\renewcommand{\baselinestretch}{0.945}

\usepackage[utf8]{inputenc} 
\usepackage[T1]{fontenc}    
\usepackage[bookmarks=false]{hyperref}

\usepackage{url}            
\usepackage{booktabs}       
\usepackage{amsfonts,amsthm,amsmath}       
\usepackage{nicefrac}       
\usepackage{microtype}      
\usepackage{xcolor}         
\usepackage{float}
\usepackage{graphicx}
\usepackage{multicol}

\usepackage{algorithm}
\usepackage{algpseudocode}
\usepackage{subcaption}

\usepackage{amsmath}
\usepackage{multirow} 

\definecolor{deepblue}{HTML}{1f77b4}
\definecolor{orange}{HTML}{ff7f0e}
\definecolor{deepgreen}{HTML}{2ca02c}
\definecolor{deepred}{HTML}{d62728}
\definecolor{grey}{HTML}{808080}
\definecolor{purple}{HTML}{800080}

\DeclareRobustCommand{\legendsquare}[1]{%
  \textcolor{#1}{\rule{1ex}{1ex}}%
}

\theoremstyle{plain}
\newtheorem{theorem}{Theorem}[section]

\theoremstyle{definition}

\newtheorem{remark}[theorem]{Remark}

\newcommand{\cblue}{\color{black}}

\newcommand{\tauh}{{\tau_h}} 
\newcommand{\gm}{{\phi^{t-1}}} 
\newcommand{\htmo}{{h_i^{t,s-1}}}
\newcommand{\taup}{{\tau_\phi}}
\newcommand{\htm}{{h_i^{t,s}}}
\newcommand{\ptm}{{\phi_i^{t,s}}}
\newcommand{\ptmo}{{\phi_i^{t,s-1}}}
\newcommand{\x}{{\bf{x}}}
\newcommand{\D}{{\mathcal{D}}}
\newcommand{\R}{{\mathbb{R}}}
\newcommand{\Y}{{\mathcal{Y}}}
\renewcommand{\H}{{\mathcal{H}}}
\newcommand{\E}{{\mathbb{E}}}
\newcommand{\GM}{{\mathrm{GM}}}
\newcommand{\Krum}{{\mathrm{Krum}}}

\newcommand{\GD}{{\mathrm{GD}}}

\makeatletter
\newcommand\fs@betterruled{%
  \def\@fs@cfont{\bfseries}\let\@fs@capt\floatc@ruled
  \def\@fs@pre{\vspace*{5pt}\hrule height.8pt depth0pt \kern2pt}%
  \def\@fs@post{\kern2pt\hrule\relax}%
  \def\@fs@mid{\kern2pt\hrule\kern2pt}%
  \let\@fs@iftopcapt\iftrue}
\floatstyle{betterruled}
\restylefloat{algorithm}
\makeatother

\begin{document}
\title{Byzantine Resilient Federated Multi-Task Representation Learning}

\author{Tuan Le and Shana Moothedath
\thanks{T. Le is with the Department of Computer Science and S. Moothedath is with the Department of Electrical and Computer Engineering, Iowa State University, Ames, IA, USA. Email: $\{$tuanle, mshana$\}$@iastate.edu.}
\thanks{This work is supported by National Science Foundation grant 2213069.}}

\maketitle
\begin{abstract}
In this paper, we propose BR-MTRL, a Byzantine-resilient multi-task representation learning framework that handles faulty or malicious agents. Our approach leverages representation learning through a shared neural network model, where all clients share fixed layers, except for a client-specific final layer. 
This structure captures shared features among clients while enabling individual adaptation, making it a promising approach for leveraging client data and computational power for personalized federated learning.
 To learn the model, we employ an alternating gradient descent strategy: each client optimizes its local model, updates its final layer, and sends estimates of the shared representation to a central server for aggregation. To defend against Byzantine agents, we employ {\cblue two robust aggregation methods for client-server communication, Geometric Median and Krum.} Our method enables personalized learning while maintaining resilience in distributed settings.
We implemented the proposed algorithm in a federated testbed built using Amazon Web Services (AWS) platform and compared its performance with various benchmark algorithms and their variations.
Through experiments using real-world datasets, including CIFAR-10 and FEMNIST, we demonstrated the effectiveness and robustness of our approach and its transferability to new unseen clients with limited data, even in the presence of Byzantine adversaries.
\end{abstract}
\begin{IEEEkeywords}
Byzantine resilience, Multi-task representation learning, Federated learning, Alternating gradient descent
\end{IEEEkeywords}
\IEEEpeerreviewmaketitle
\section{Introduction}
Machine learning has traditionally relied on centralized settings, where a single model is trained on large datasets stored in one central location. This approach requires gathering data from all clients into a central data center, which raises significant data privacy concerns. Federated Learning (FL) addresses these challenges by using a distributed framework that preserves data privacy while overcoming the problem of isolated data sources \cite{mcmahan2017communication}. FL enables multiple clients, such as mobile devices or institutions, to collaboratively train a global model without sharing their raw data. A central server is responsible for coordinating the clients by aggregating the models received from each client and broadcasting the updated model back to them in each round of communication. McMahan {\em et al.} introduced the Federated Averaging (FedAvg) algorithm, which facilitates the training of a single shared global model designed to achieve optimal performance on average across all participating clients \cite{mcmahan2017communication}. The approach offers several advantages, such as simplicity in implementation and a streamlined aggregation process, making it scalable for large client populations. Additionally, the global model promotes uniformity, ensuring consistent performance for clients with similar data distributions. However, FL approaches encounter three primary bottlenecks: (i) {\em Personalization:} In real-world scenarios, collaborating client tasks are typically `different yet related', making a single global model suboptimal for all tasks. Instead, personalized models are better suited to capture the unique characteristics and requirements of each client. (ii) {\em Transferability:} Federated models often struggle to generalize to unseen clients that were not part of the training process, limiting their effectiveness in dynamic, real-world applications. (iii) {\em Byzantine attacks:} FL is particularly vulnerable to Byzantine attacks, as it relies on distributed client updates; a faulty or malicious client can potentially corrupt the global model, compromising performance and reliability.

Recent works proposed personalized FL algorithms. 
Deng {\em et al.} proposed the Adaptive Personalized Federated Learning (APFL) algorithm by combining local and global model optimization \cite{deng2020adaptive}. This method allows clients to train personalized models that better generalize on local distributions while still contributing to the learning of a robust global model. Fallah {\em et al.} proposed a method called Personalized Federated Learning via Model-Agnostic Meta-Learning (MAML) by applying a meta-learning framework \cite{fallah2020personalized}. This method allows clients to quickly adapt a shared initial model to their local data distributions by performing a few steps of gradient descent, leading to personalized models for each user. 

To achieve generalizability, Multi-Task Representation Learning (MTRL) has emerged as a promising methodology, aiming to address multiple related tasks/clients simultaneously by leveraging shared representations \cite{bengio2013representation}. 
The core principle of MTRL is based on the idea that, even though data in federated settings is often non-i.i.d. (non-independent
and identically distributed) across clients, there exists a shared learned representation across multiple clients that effectively captures the common structural elements inherent to each client. This shared foundation enables the use of simple linear predictors in the final model layer, which are then fine-tuned to address the unique variations and requirements of each client. This approach significantly enhances the efficiency of model training processes, as it allows for the transfer and utilization of learned features across  {\em different but related tasks}. MTRL has demonstrated remarkable success across various applications such as natural language processing domains GPT-2, GPT-3, Bert,  vision
domains CLIP, and robotics.
Recent studies \cite{collins2021exploiting, du2020few, OurICML, lin2024fast} have focused on the theoretical understanding of MTRL, particularly when the shared model exhibits specific structures, such as a low-rank representation. 

To address Byzantine attacks in federated learning, Gouissem {\em et al.} developed a collaborative cross-check mechanism by allowing clients to collaboratively identify malicious nodes based on each other’s contributions to model updates \cite{gouissem2023collaborative}. While effective, this approach introduces significant computational overhead due to the extensive cross-validation among clients. In contrast, Lin {\em et al.} tackled Byzantine resilience using robust aggregation techniques, such as geometric median, Krum, and h-Krum, to address the matrix completion problem \cite{lin2019byzantine}. Recently, Li {\em et al.} studied Byzantine resilient FL for non-iid data by obtaining an inversed artificial gradient \cite{li2024byzantine}. Zhang {\em et al.} investigated communication-efficient approaches for Byzantine-resilient FL and proposed a projected stochastic gradient descent (SGD) algorithm to enhance system robustness \cite{zhang2024byzantine}.
Their approach employed a Huber function-based robust aggregation with an adaptive threshold selecting
strategy at the server to reduce the effects of Byzantine attacks.

Despite these advancements, a unified approach for Byzantine-resilient personalized FL with transferability remains underexplored. In this paper, we introduce the Byzantine-resilient Multi-Task Representation Learning (BR-MTRL) framework to jointly address three critical challenges in FL: personalization, transferability, and Byzantine resilience. By incorporating geometric median {\cblue and Krum} aggregation methods into the shared representation, while updating the task-specific parameter locally we enhance the global model’s resilience against Byzantine attacks. Our framework ensures that the models learned across distributed agents remain robust, reliable, and secure, even in the presence of adversarial influences, improving both performance and safety.

\vspace{-1 mm}
\subsection{Contributions}
The contributions of this paper are threefold. 

\begin{itemize}
    \item We propose a BR-MTRL algorithm for personalized federated representation learning. In our approach, all clients share a common representation while maintaining their local personalized heads. We introduce an alternating gradient descent approach that alternates between solving the shared model and fine-tuning the task-specific head of each client locally. To ensure robustness, our method employs Geometric Median (GM) and {\cblue Krum}-based mechanisms to aggregate client updates on the server, allowing clients to learn a Byzantine-resilient shared representation while preserving task-specific personalizations.
    \item We implemented our proposed approach in a truly federated environment by deploying it on a simulation testbed built on the Amazon Web Services (AWS) platform. We performed simulations on datasets, including CIFAR-10 and FEMNIST on the AWS testbed, and validated the effectiveness of BR-MTRL in mitigating the impact of Byzantine attacks while maintaining learning performance.
    \item We conducted a meta-test to evaluate the transferability of the learned representation on new, unseen clients with limited data. The BR-MTRL approach was assessed by fixing the shared representation model and learning only the personalized heads of the new clients. We measured the impact of Byzantine clients on new clients' performance and demonstrated improved accuracy using BR-MTRL, ensuring the transferability of our approach.
\end{itemize}

\vspace{-1 mm}
\subsection{Related Work}
Personalized Federated Learning (PFL) has received a lot of research interest in recent years. Liang {\em et al.}  \cite{liang2020think} proposed the Local Global Federated Averaging (LG-FedAvg) method, which focuses on learning low-dimensional features from clients' data. 
Their approach emphasizes learning multiple local representations and a single global head. On the other hand, Arivazhagan {\em et al.} \cite{arivazhagan2019federated} proposed FedPer algorithm that explicitly divides the model architecture into base layers, trained across all devices via federated averaging and personalization layers, which are tailored locally on each device using its specific data. This structure not only promotes personalization by adapting the model to local data characteristics but also maintains a cohesive model performance across diverse environments by leveraging a shared base learning across all clients. In addition, Collins {\em et al.} \cite{collins2021exploiting} proposed the FedRep algorithm focusing on learning a global model shared across clients while allowing each client to maintain a personalized head model, similar to \cite{arivazhagan2019federated}. 
The algorithms in \cite{collins2021exploiting,arivazhagan2019federated} allow clients to collaborate on learning shared representations while maintaining personalized models tailored to their specific data distribution, similar to our approach; however, they focus on PFL without considering the challenges posed by Byzantine attacks.

To counter Byzantine attacks, many PFL systems incorporate robust aggregation techniques at the server, such as the geometric median (GM) and its variations. For example, Li \emph{et al.} introduced AutoGM, an auto-weighted GM aggregation rule designed to enhance robustness in the presence of Byzantine failures, specifically model and data poisoning attacks \cite{li2021byzantine}. AutoGM improves the classic GM by automatically excluding extreme outliers and re-weighting the remaining parameter updates based on a skewness threshold, offering a more adaptive and flexible solution. Similarly, Wang \emph{et al.} proposed WGM-dSAGA, a weighted GM for robust aggregation \cite{wang2023wgm}. The strategy incorporates COPOD, a parameter-free outlier detection technique, to identify and assign weights to client updates based on their potential maliciousness. 
Meanwhile, He \emph{et al.} tackled Byzantine attacks by introducing a novel projected stochastic block gradient descent method \cite{he2023byzantine}. Unlike the GM-based approaches, their strategy employs a server aggregation technique that combines local models using a Huber function-based descent step to manage the impact of malicious updates from Byzantine clients. While these studies have made significant strides in mitigating Byzantine attacks, they primarily focus on enhancing aggregation techniques on the server side. In contrast, our work introduces a Byzantine-resilient MTRL framework that learns and leverages a shared representation to enhance the quality of each client's model, while ensuring both transferability and resilience.
\vspace{-1mm}
\section{Problem Setting}
We address a multi-task supervised learning scenario with $n$ clients, where each $i$-th client possesses $M_i$ labeled samples $\{\x_i^j, y_i^j\}_{j=1}^{M_i}$, generated from a distribution $\D_i$. The goal is to learn a model $q_i: \R^d \rightarrow \Y$ maps input $\x_i \in \R^d$ to predicted labels $q_i(\x_i)$, such that $q_i(\x_i)\in\Y$ is 
as close as possible to the true label $y_i$. That is, to minimize the expected risk over $\D_i$ given by $f_i:=\E_{(\x_i, y_i)\sim \D_i} [\ell(q(\x_i), y_i)]$, where $\ell: \Y \times \Y \rightarrow \R$ is a loss function that models the error between $q(\x_i)$ and $y_i$. 

In FL paradigms, $M_i$'s are typically small, and if clients train a model using their local data separately, it may lead to poor generalization and performance due to limited data. 
To this end, agents collaborate by sharing model updates instead of raw data, leveraging the collective knowledge of all clients while preserving data privacy. Federated learning enables this cooperation through a central server, allowing clients to learn models using aggregated data while maintaining privacy. The standard FL approach thus learns a global model $q=q_1=q_2=\ldots=q_n$ by minimizing the average error $1/n\sum_{i=1}^n f_i(q)$ \cite{mcmahan2017communication,collins2021exploiting}. Unlike the standard setting, we focus on personalized models and learning the shared representations.

We consider a model comprising a global representation, $\phi: \mathbb{R}^d \rightarrow \mathbb{R}^k$, and a local model for the $i$-th client, $h_i: \mathbb{R}^k \rightarrow \Y$. Specifically, the model for each client is defined as the composition of these two mappings, expressed as $q_i(\x_i) = (h_i \circ \phi)(\x_i)$. We assume that the dimensionality of the local model, $k \ll d$, meaning the number of parameters to be learned at the local client is much smaller than the overall model size. This captures the relationships among clients and enables model transferability in data-scarce scenarios, as only a small number of parameters are fine-tuned for new clients, while the global shared model $\phi$ remains fixed.
The objective function of our learning process is thus formulated as
\vspace{-2 mm}
\begin{align}\label{eq:loss}
   \min_{\phi \in \Phi} \frac{1}{n} \sum_{i=1}^n \min_{h_i \in \H} f_i(h_i \circ \phi), 
\end{align}
where $\Phi$ and $\H$ represent the classes of possible global and local representations, respectively.
\subsection{System Model}
\begin{figure}
    \centering
    \includegraphics[width=0.98\linewidth]{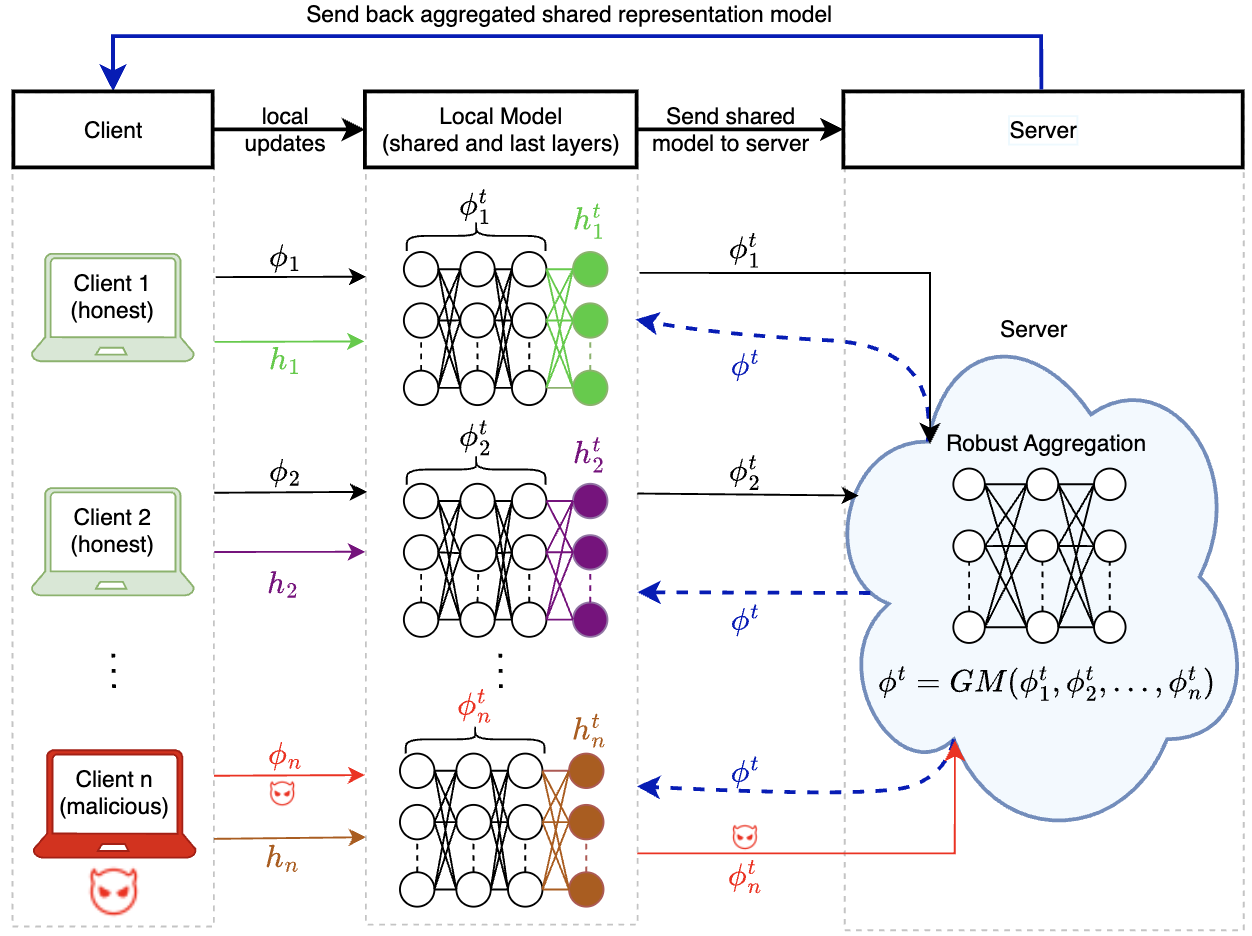}
    \vspace{-2 mm}
    \caption{The system model of proposed MTRL.}
    \label{fig: system}
    \vspace{-4 mm}
\end{figure}

Our scheme comprises $n$ clients and a central server, with a subset of clients being malicious.
The system model of the proposed MTRL scheme is presented in Figure~\ref{fig: system}.
In each iteration round $t$ the proposed scheme follows these steps: (1) Each client fine-tunes and updates its local head $h^t_i$ using local data. (2) Clients update their estimates of the shared representation $\phi^t_i$. (3) Clients upload their updates to the server. (4) The server applies geometric median or Krum-based robust aggregation to compute the new shared representation $\phi^t$. (5) The updated representation is sent back to all clients.

\noindent{\bf Malicious clients.} We consider two types of clients: Byzantine and honest. Byzantine clients attack the global model update in two different ways: 1) corrupt the model sent to the central server by providing malicious model updates and 2)~provide incorrect labels for their data. The attacker's goal is to corrupt the final learned model. In contrast, honest clients contribute genuine updates, ensuring reliable and accurate training. 

\noindent{\bf Robust server.} To minimize the impact of Byzantine clients, we apply the Geometric Median (GM) and Krum to the weights received from clients, which is then transmitted to the server. The aggregation at the server is for the shared model among all the clients, which motivates the  selection of GM and Krum aggregation techniques.
By using the median instead of the mean, GM reduces outlier distortion, ensuring the global model reflects honest client updates. Krum enhances resilience by selecting the client update that is most similar to its nearest neighbors, thereby mitigating the impact of malicious updates.

\vspace{-1 mm}
\subsection{Design Goals}
In order to solve the MTRL problem under the above threats, our design should
achieve the following goals.

\noindent{\bf Representation learning and model accuracy.} The proposed method must jointly learn a shared representation model and personalized (local) client/task specific heads, while ensuring acceptable accuracy across the tasks.

\noindent{\bf Transferability.} The proposed scheme should generalize to new, unseen clients/tasks by leveraging the learned shared model while training only the local client head, enabling efficient meta learning even with limited data.

\noindent{\bf Byzantine robustness.} The proposed scheme learns an unbiased, personalized model, which minimizes the impact of malicious model updates and mislabeled data.
\section{Proposed Byzantine Resilient MTRL Algorithm}
We propose an alternating gradient descent (GD) approach, the pseudocode of which is presented in Algorithm~\ref{alg:byz}. The goal of our algorithm is to jointly learn the global model $\phi$ and to learn the local client heads $h_i$, for $i\in [n]$, locally to minimize Eq.~\eqref{eq:loss}. The algorithm begins with the central server initializing the shared model as $\phi^0$ randomly.
Then, each round of the algorithm involves two main steps: (i) Client update via GD to update $h_i$ after fixing $\phi$ and (ii)~Server update via GD at the client side to update $\phi_i$ by fixing $h_i$, followed by robust aggregation at the server. We elaborate each of the steps below.
%

\textbf{Client Update.} In each round $t \in [T]$, all clients participate in the client update. During this update, client $i$ performs $\tauh$ local gradient steps to optimize its head model based on the current global representation $\gm$, which is shared by the server. Specifically, for each step $s = 1, \dots, \tauh$, client $i$ updates its head model as follows:
\[
\htm = \GD(f_i(\htmo, \gm), \htmo, \eta).
\]

$\GD(f, h, \eta)$ represents a generic gradient descent update of the variable $h$ using the gradient of the function $f$ with respect to $h$, and the step size $\eta$. Different gradient descent methods can be used, such as SGD or SGD with momentum. Each client makes many such local updates, i.e., $\tauh$ is large, to optimize its local head based on the most recent global representation $\gm$ received from the server, leveraging its own local data.
Next, the client performs $\taup$ local updates to refine its estimate of shared representation $\phi_i$ after fixing the local head estimate $h_i^{t, \tauh}$ as, for $s = 1, \dots, \taup$,
\[
\ptm = \GD(f_i(h_i^{t, \tauh}, \ptmo),\ptmo, \eta).
\]

\textbf{Server Update.} Once the local updates for the head and shared representation are completed, the client updates the server by sending its locally updated representation denoted by $\phi_i^{t,\tau_\phi}$ to the server. We consider two types of Byzantine behaviors among malicious clients. In the \emph{Scaled Random (SR)} attack, Byzantine clients inject scaled random noise into their local model updates to distort the global aggregation. 
\[
\phi_i^{t,\taup} = \phi_i^{t,\taup} + \sigma \cdot \mathcal{N}(0, I),
\]
where $\sigma$ is a scalar and $\mathcal{N}(0, I)$ denotes standard Gaussian noise. In the \emph{Mislabeling (ML)} attack, Byzantine clients switch the labels of their local training data to cause misclassification and misleading gradients. All clients $i\in [n]$ sends their local estimate of the representation, $\phi_i^{t,\taup}$, to the central server. 

\textbf{Robust Aggregation.} To ensure robust aggregation, the server calculates GM for each layer of the shared model by finding a matrix $P$ by solving the following optimization
\[
\vspace{-2 mm}
\GM(\phi_1^{t,\tau_\phi}, \cdots, \phi_n^{t,\tau_\phi}) = \arg \min_{P\in\mathbb{R}^{d \times k}} \sum_{i=1}^{n}\|\phi_i^{t,\tau_\phi} - P\|_F,
\]
where $\|\cdot\|_F$ denote Frobenius norm. The server sets the GM as the estimate of the representation $\phi^t$ and sends it to all clients. The three steps are repeated for $T$ iterations.

{\cblue Further, we perform a Krum approach \cite{lin2019byzantine} detailed below. 
\[
\Krum(\phi_1^{t,\tau_\phi}, \cdots, \phi_n^{t,\tau_\phi}) = \phi_{k^*}^{t,\tau_\phi}, \mbox{~where}
\]
\[
 k^* = \arg \min_{k \in [n]} \sum_{k\rightarrow k'}\|\phi_k^{t,\tau_\phi} - \phi_{k'}^{t,\tau_\phi}\|_F.
\]
For $k\neq k'$, $k \rightarrow k'$ is the set of $n-f-2$ nearest neighbors of $\phi_k^{t,\tau_\phi}$, where $f$ is the number of Byzantine clients. Thus $\phi_{k^*}^{t,\tau_\phi}$ is the model with minimal summed distance to its neighbors. 
\vspace{-2 mm}
\begin{remark}
Krum requires to know the number of Byzantine clients in advance, and its computational complexity scales quadratically with the number of clients, resulting in significantly higher computational requirements as compared to GM.
\end{remark}
}

\begin{algorithm}[h]
\caption{Byzantine-Resilient Multi-Task Representation Learning (BR-MTRL) Algorithm}\label{alg:byz}
\textbf{Parameters:} step size $\eta$, number of local updates for the head $\tau_h$ and for the global representation $\tau_\phi$, number of communication rounds $T$. \\
\textbf{Initialize:} $\phi^0, h_1^{0,\tau_h}, \dots, h_n^{0,\tau_h}$
\begin{algorithmic}[1]
\For{$t = 1,2,  \dots, T$}
    \State Server sends current representation $\phi^{t-1}$ to all clients
    \For{each client $i = 1,2, \dots, n$}
        \State Client $i$ initializes $h_i^{t,0}$, $h_i^{t,0} \gets h_i^{t-1, \tau_h}$ 
        \For{$s = 1$ to $\tau_h$}\Comment{{\footnotesize Client $i$ makes $\tau_h$ head updates}}
            \State $h_i^{t,s} \gets \text{GD}(f_i(h_i^{t,s-1}, \phi^{t-1}), h_i^{t,s-1}, \eta)$
        \EndFor
        \State Client $i$ initializes $\phi_i^{t,0}$ $\phi_i^{t,0} \gets \phi^{t-1}$ 
        \For{$s = 1$ to $\tau_\phi$}\Comment{{\footnotesize Client $i$ makes $\tau_\phi$ updates to $\phi$}}
            \State $\phi_i^{t,s} \gets \text{GD}(f_i(h_i^{t,\tau_h}, \phi_i^{t,s-1}), \phi_i^{t,s-1}, \eta)$
        \EndFor
        
        \If{Client $i$ is malicious}
            \State $\phi_i^{t,\tau_\phi} = Attack(\phi_i^{t,\tau_\phi})$\label{line:byz}
        \EndIf 
        
        \State Client $i$ sends updated representation $\phi_i^{t,\tau_\phi}$ to server
    \EndFor
    \State Server computes the new representation as
    $\phi^t \gets \mathrm{GM}(\phi_1^{t,\tau_\phi}, \cdots, \phi_n^{t,\tau_\phi}) \mbox{~or~} \mathrm{Krum}(\phi_1^{t,\tau_\phi}, \cdots, \phi_n^{t,\tau_\phi})$
\EndFor
\end{algorithmic}
\end{algorithm}

\vspace{-2 mm}
\section{Experimental Analysis}
\subsection{Experimental Setup}
\noindent{\bf Datasets.} We evaluated our algorithm on two real-world datasets, CIFAR-10 and FEMNIST. CIFAR-10 consists of 60,000 $32\times32$ color images in 10 classes, with 50,000 training images and 10,000 test images.
FEMNIST consists of 145,600 $28\times 28$ pixel images of 26 handwritten characters, with 124,800 training images and 20,800 test images.

\noindent{\bf Model.} We use a deep neural network, which is composed of two convolutional blocks followed by a linear layer for the shared representation and an additional linear layer for the personalized client heads. Each convolutional block includes a convolutional layer with a stride and padding of $1$, a ReLU activation function, and a max-pooling layer with a kernel size of $2$. We optimize the model using stochastic gradient descent (SGD) with a learning rate $\eta = 0.01$ and momentum $\beta = 0.9$. 

\noindent{\bf Hyperparameters.} We initialized all models with random parameters. For CIFAR-10, training was conducted for $T = 100$ communication rounds, while FEMNIST models were trained for $T = 200$ rounds. In each round, all the clients perform local updates and receive the aggregated model from the server. Each client executed 10 local epochs of SGD with momentum for head training and one epoch for updating the global representation.

The data distribution followed a highly non-i.i.d., pathological partitioning scheme, where each client was assigned a fixed subset of classes and, for each class, randomly sampled 750 images. We considered $100$ and $1000$ clients for CIFAR-10 and $150$ clients for FEMNIST. Among them, $20$ out of $100$, $100$ out of $1000$, and $50$ out of $150$ clients were designated as malicious. By varying the proportion of Byzantine clients, we analyzed their impact on the learned model.

\noindent{\bf Benchmark Algorithms.}
We evaluated our algorithm with benchmark approaches, including FedAvg \cite{mcmahan2017communication}, FedPer \cite{arivazhagan2019federated}, and FedRep \cite{collins2021exploiting}. FedPer and FedRep are similar approaches that learn a shared representation with personalized heads; FedPer updates both $\phi, h$ simultaneously within each round $t$ while FedRep updates $\phi, h$  in an alternating fashion. In contrast, FedAvg learns a single global model without personalization.
As a best-case benchmark, we considered these methods without malicious clients. Additionally, we analyzed these variants: (i) introducing malicious clients by adding noise to local representation (SR FedPer/FedRep/FedAvg), (ii) introducing malicious clients by misclassifying the image label (ML FedPer/FedRep/FedAvg), (iii) incorporating GM aggregation (FedAvg+GM, FedPer+GM), and (iv) incorporating Krum aggregation (FedAvg+Krum, FedPer+Krum). For meta-test, we consider an additional baseline, Naive, which trains new clients independently without leveraging the MTRL framework.

\noindent{\bf Performance metrics.} We evaluate the performance of our algorithm by calculating the average classification accuracy across all benign clients in the presence of malicious clients.

\subsection{Amazon Web Services (AWS)-Based Simulation Testbed}
%
We deployed our proposed approach on AWS using two experimental setups. In the first setup, called {\em sequential AWS}, a single EC2 (Elastic Compute Cloud) instance is used for the server, while another EC2 instance hosts all the clients. In the second setup, {\em federated AWS}, each client is allocated a separate EC2 instance. In both setups, during each communication round, the server retrieves the shared model from S3 (Simple Storage Service) cloud storage and sends it along with the selected clients’ IDs to the clients. In the sequential AWS setup, clients update their respective heads sequentially and send the updated model back to the server. In the federated AWS setup, the server sends the shared model and client IDs to the respective EC2 instances, where each client performs local learning, updates its head, and sends the updated model back to the server in parallel. Thus, sequential AWS mimics a standard simulation of an FL algorithm, while federated AWS represents a genuinely federated simulation platform. We compared the communication and computation times for both setups and evaluated their performance.

We considered a synchronous broadcast model, where the server waits until it receives the updated models from all selected clients before aggregation.
After updating its head, the client stores the new model in its S3 bucket. We deployed the system on an AWS EC2 t2.large instance (2 vCPUs, 8 GiB RAM, 64 GB EBS storage). Communication between the server and clients is handled via Flask, which enables HTTP requests for model updates. The server sends requests to clients, and Flask processes like receiving the shared model estimates and sending back updated models. This setup ensures efficient communication and aggregation of client models. 
\begin{table}[t]
\vspace{4 mm}
    \renewcommand{\arraystretch}{1.3} 
    \setlength{\tabcolsep}{5pt} 
    \begin{tabular}{|c|c|c|}
        \hline
        \textbf{Process} & \textbf{Federated AWS} & \textbf{Sequential AWS} \\
        \hline
        Sending from server to clients & 0.31 & 0.13 \\
        Training clients' model & 15.41 & 92.79 \\
        Sending from clients to server & 4.68 & 1.20 \\
        Average time cost per round & 26.50 & 121.75 \\
        \hline
    \end{tabular}
    \caption{Comparison of average runtime (in milliseconds) for federated AWS and sequential AWS implementations.}
    \label{tab:runtime_comparison}
    \vspace{-4 mm}
\end{table}
We present the running time comparison in Table \ref{tab:runtime_comparison}. Training time in Table \ref{tab:runtime_comparison} represents the total time for all clients to train their models. The results show that federated AWS achieves significantly faster communication rounds than sequential AWS due to parallel client training, which greatly reduces overall training time. Although federated AWS incurs slightly higher communication overhead between the server and clients, the overall efficiency gains make FL an effective approach.
\vspace{-2 mm}
\subsection{Experiment Results}
\begin{figure*}[t]
\subcaptionbox{\label{fig:cfsr} CIFAR-10 (SR)}{\includegraphics[scale=0.225]{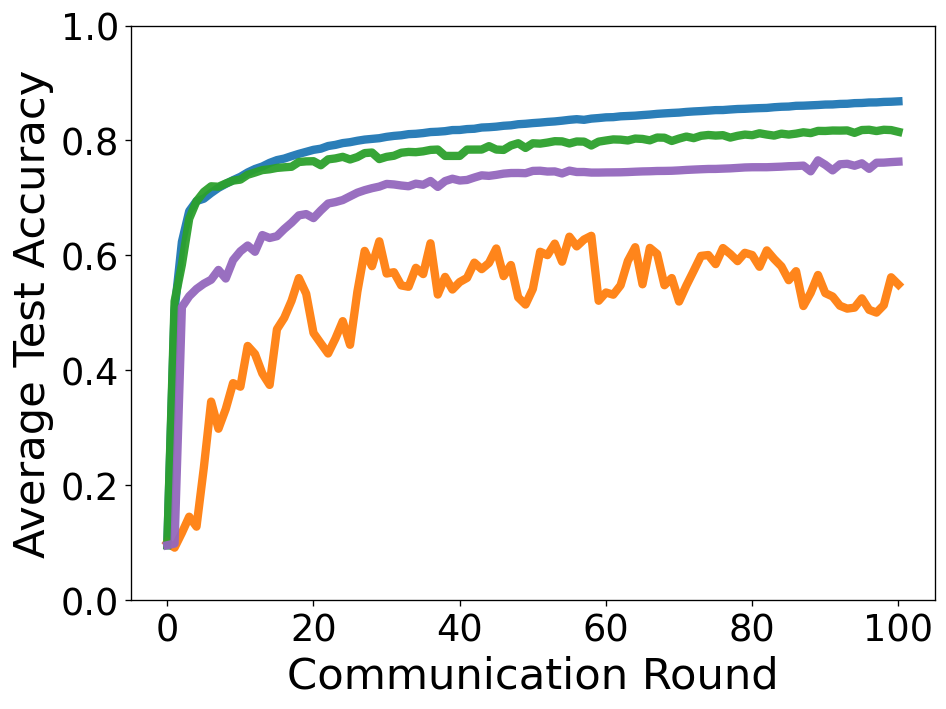}}
\hspace{-0.25 em}%
\subcaptionbox{\label{fig:emsr} FEMNIST (SR)}{\includegraphics[scale=0.225]{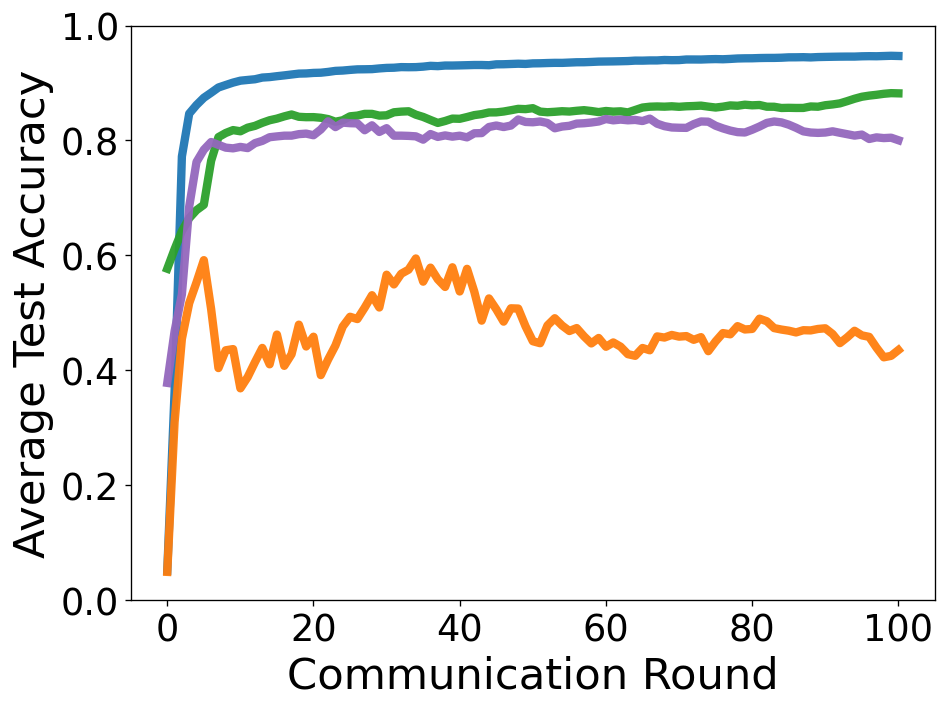}}
\hspace{-0.25 em}%
\subcaptionbox{\label{fig:metacfsr} Meta test: CIFAR-10 (SR)}{\includegraphics[scale=0.225]{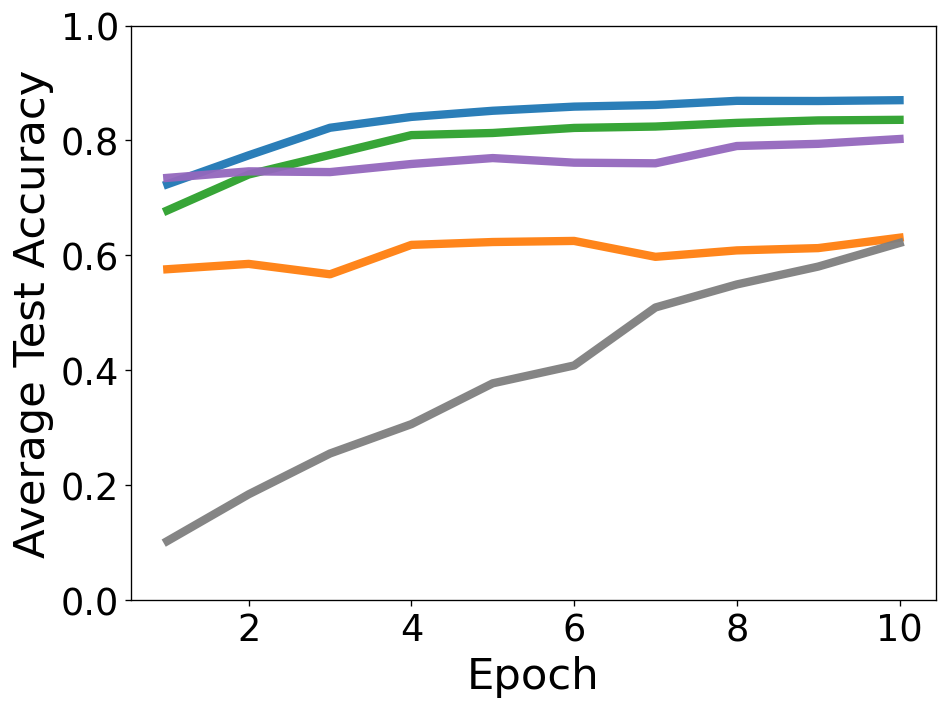}}
\hspace{-0.25 em}%
\subcaptionbox{\label{fig:metaemsr} Meta test: FEMNIST (SR)}{\includegraphics[scale=0.225]{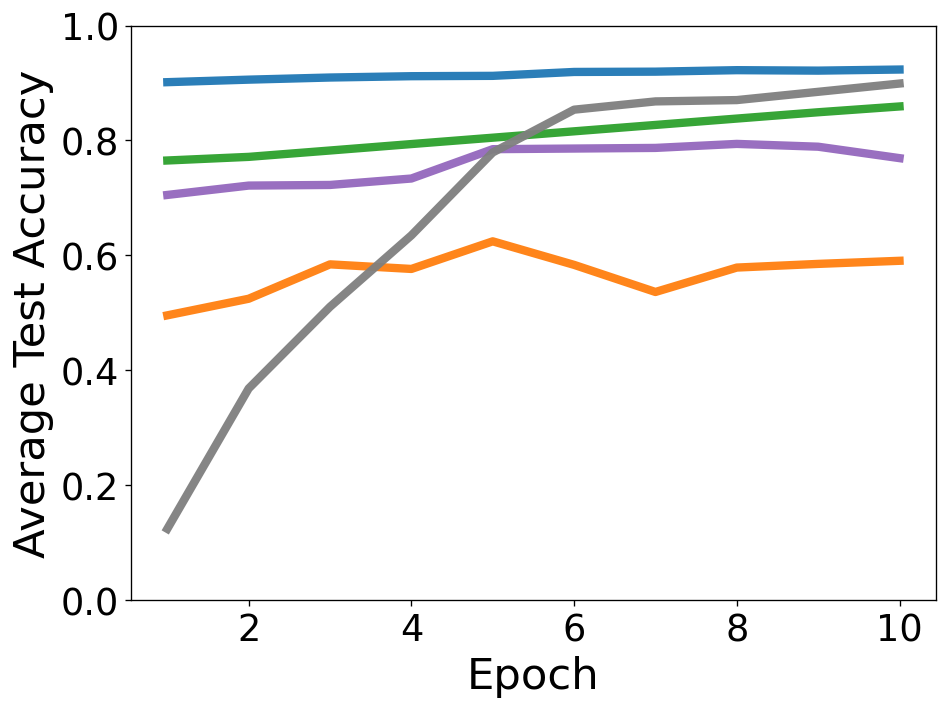}}

\subcaptionbox{\label{fig:cfml} CIFAR-10 (ML)}{\includegraphics[scale=0.225]{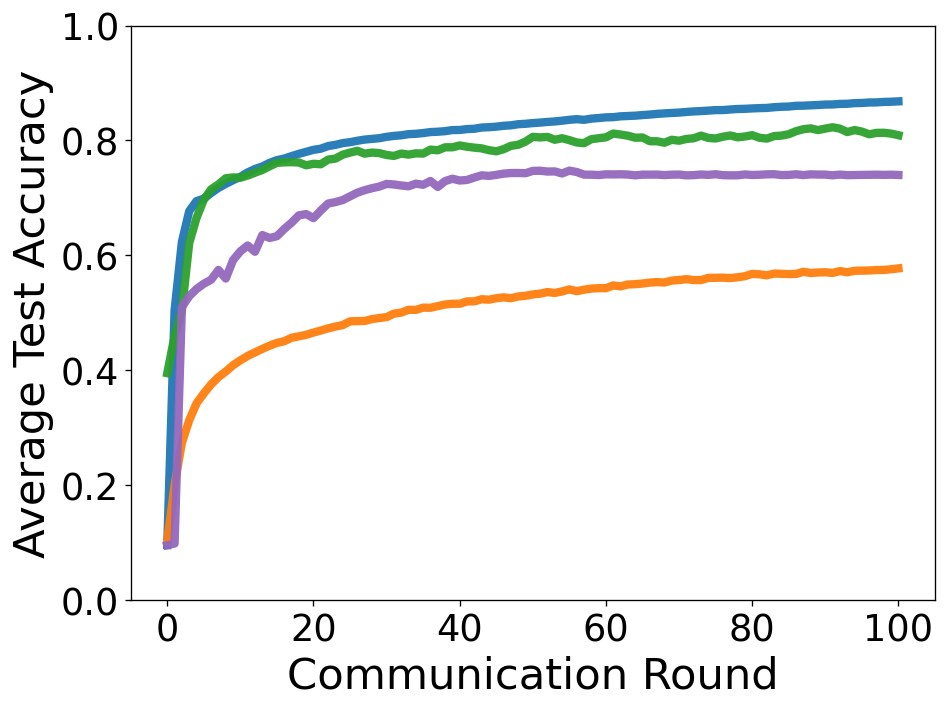}}
\hspace{-0.25 em}%
\subcaptionbox{\label{fig:emml}FEMNIST (ML)}{\includegraphics[scale=0.225]{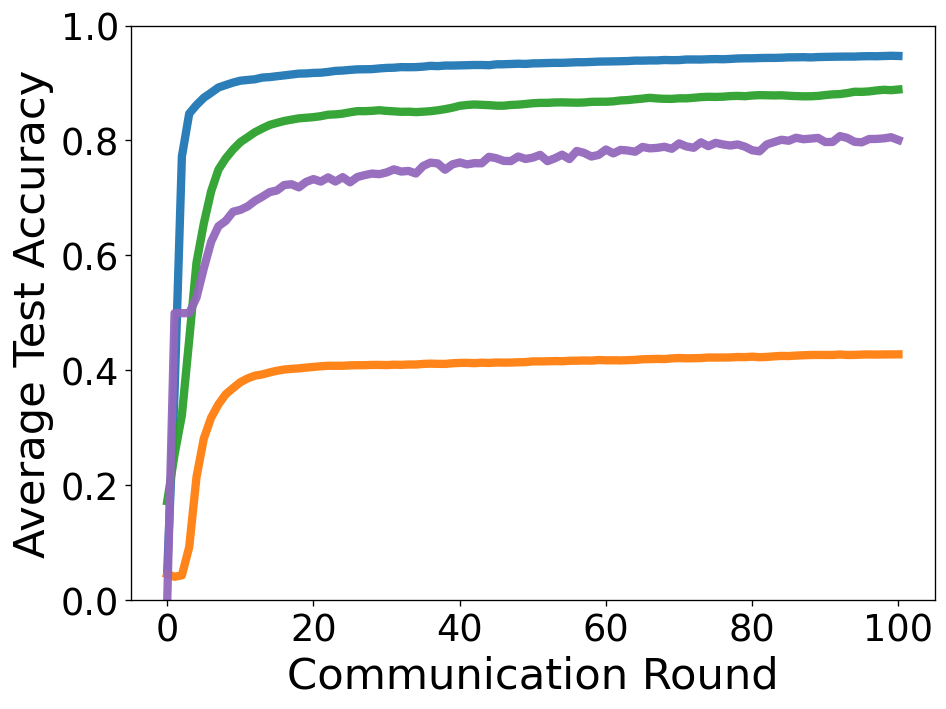}}
\hspace{-0.25 em}%
\subcaptionbox{\label{fig:metasr} Meta test: CIFAR-10 (ML)}{\includegraphics[scale=0.225]{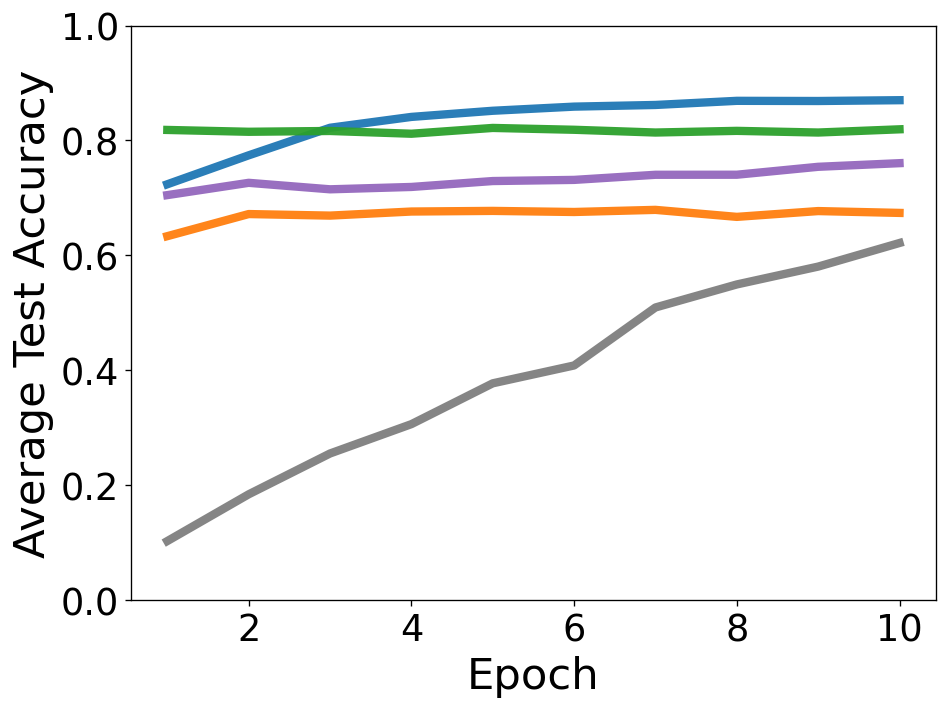}}
\hspace{-0.25 em}%
\subcaptionbox{\label{fig:metaml} Meta test: FEMNIST (ML)}{\includegraphics[scale=0.225]{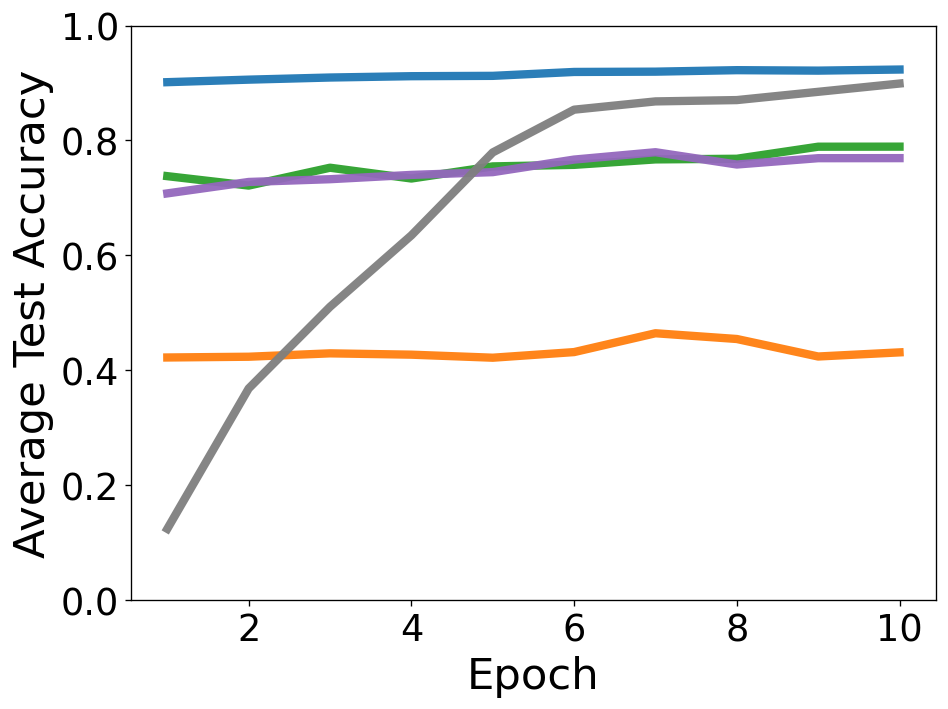}}
\vspace{-2 mm}
\caption{\small \legendsquare{deepblue}~FedRep,
  \legendsquare{orange}~Byzantine FedRep,
  \legendsquare{deepgreen}~BR-MTRL+GM (proposed), \legendsquare{purple}~{\cblue BR-MTRL+Krum}  (proposed) \legendsquare{grey}~Naive.  {\bf Average test accuracy across all clients:} Figure~\ref{fig:cfsr} and~\ref{fig:cfml} show the results for CIFAR-10 with 100 clients with 20 malicious clients under SR and ML attacks respectively. Figure~\ref{fig:emsr} and ~\ref{fig:emml} present the results for FEMNIST with 150 clients with 50 malicious clients under SR and ML attacks respectively. We compared our proposed BR-MTRL with the FedRep algorithm under benign conditions (i.e., no malicious client) and FedRep subjected to Byzantine attacks (Byzantine FedRep). {\bf Meta test for new clients:}  Figures~\ref{fig:metacfsr}, \ref{fig:metaemsr}, \ref{fig:metaml} and~\ref{fig:metasr} present the average test accuracy across 10 new clients (CIFAR-10 and EMNIST) after fine-tuning the local heads for 10 epochs using the shared model. We first learned the shared model using source clients and then used the shared model to optimize the local head of new clients.}\label{fig:main1}
  \vspace{-3 mm}
\end{figure*}
\begin{table}[t]
\vspace{4 mm}
\centering
\renewcommand{\arraystretch}{1}
\setlength{\tabcolsep}{3pt}
\begin{tabular}{p{2.7cm}|p{1.2cm}p{1.2cm}p{1.5cm}|p{1.2cm}}
\hline
\multirow{2}{*}{\textbf{Method}} & \multicolumn{3}{c|}{\textbf{CIFAR-10}} & \textbf{FEMNIST} \\ \cline{2-5}
 & (100,2,20) & (100,5,20) & (1000,2,100) & (150,3,50) \\ \hline
\textbf{FedAvg}~\cite{mcmahan2017communication}  & 38.56 & 55.39 & 60.81 & 61.46 \\
\textbf{FedPer}~\cite{arivazhagan2019federated}  & 82.03 & 75.59 & 75.10 & 93.77 \\
\textbf{FedRep}~\cite{collins2021exploiting}     & 86.78 & 82.11 & 81.11 & 94.69 \\ \hline \hline
\textbf{SR FedAvg}   & 10.43 & 15.52 & 11.71 & 5.78 \\
\textbf{SR FedPer}   & 52.44 & 38.51 & 54.00 & 37.37 \\
\textbf{SR FedRep}   & 54.88 & 35.41 & 54.26 & 43.47 \\ \hline
\textbf{SR FedAvg+Krum} & 15.78 & 39.30 & 32.28 & 11.50 \\
\textbf{SR FedAvg+GM}   & 38.11 & 55.18 & 35.45 & 11.45 \\
\textbf{SR FedPer+Krum} & 74.72 & 67.48 & 70.05 & 83.11 \\
\textbf{SR FedPer+GM}   & 76.26 & 70.12 & 73.16 & 85.73 \\
\textbf{SR BR-MTRL+Krum} & 76.28 & 74.21 &74.68 & 80.28 \\
\textbf{SR BR-MTRL+GM}   & \textbf{81.48} & \textbf{82.03} & \textbf{75.36} & \textbf{88.47} \\ \hline \hline\textbf{ML FedAvg}   & 27.90 & 28.25 & 13.86 & 32.65 \\
\textbf{ML FedPer}   & 28.07 & 46.66 & 48.83 & 60.95 \\
\textbf{ML FedRep}   & 57.54 & 58.04 & 60.31 & 42.72 \\ \hline
\textbf{ML FedAvg+Krum}  & 31.23 & 36.02 & 45.88 & 48.16 \\
\textbf{ML FedAvg+GM}  & 30.86 & 51.67 & 52.40 & 54.91 \\
\textbf{ML FedPer+Krum}  & 71.65 & 63.15 & 68.80 & 72.00 \\
\textbf{ML FedPer+GM}  & 74.85 & 69.53 & 68.94 & 73.41 \\
\textbf{ML BR-MTRL+Krum} & 73.96 & 70.01 & 74.25 & 80.05 \\ 
\textbf{ML BR-MTRL+GM} & \textbf{79.97} & \textbf{76.15} & \textbf{80.05} & \textbf{89.11} \\ \hline
\end{tabular}
\caption{Average test accuracy (\%) across all benign clients under \textbf{Scaled Random (SR)} and \textbf{Mislabeling (ML)} Byzantine attacks for different algorithms. $(n,S,b)$ denote number of clients, classes per client, and number of Byzantine clients.}
\label{tab:SR_attack}
\vspace{-3mm}
\end{table}
%

\noindent{\bf Test accuracy.} As shown in Figures~\ref{fig:cfsr}, \ref{fig:emsr}, \ref{fig:cfml}, \ref{fig:emml} and Table~\ref{tab:SR_attack}, the FedRep and FedPer algorithms consistently achieved higher accuracy than FedAvg for all experiments, demonstrating the effectiveness of personalized FL. However, the performance of these three baselines (without malicious clients) significantly declined under Byzantine attacks. This decline validates that  (personalized) FL frameworks susceptible to Byzantine attacks will lead to unreliable models with poor performance.
The SR setting induces a more significant decline, especially as the number of classes per client increases from $(100,2,20)$ to $(100,5,20)$, indicating that greater task heterogeneity amplifies the propagation of corrupted updates. In contrast, the ML setting exhibits an increase, suggesting that personalized representations partially mitigate label-based corruption. Additionally, scalability enhances robustness across both scenarios: when the proportion of malicious clients decreases from $20\%$ $(100,2,20)$  to $10\%$ $(1000,2,100)$, the aggregation process benefits from the dominance of benign contributions.
Our experimental results further demonstrate that the proposed approaches, BR-MTRL+GM, achieve a higher accuracy than both the GM and Krum-adapted versions of FedPer. 
This is because the proposed alternating gradient descent approach within each round is more effective than FedPer’s simultaneous updates of the client head $h$ and server model $\phi$.
All experimental results consistently showcase the improved performance of the proposed approach over the two attacks. Furthermore, GM outperforms Krum in most of the experiments. One potential reason is that Krum's output is overly conservative, as it is exactly one of the client updates, discarding information from other honest clients. The GM-based solution is more effective, especially when the number of malicious clients is less. Further, Krum requires knowing the number of malicious clients, and has higher computation.

\noindent{\bf Results for Meta-test.}
We tested the transferability of the proposed approach. We first learn the shared model $\phi$ using the source clients and then utilize this model to train the local heads of new, previously unseen clients. 
 For each new client, we fine-tune the client head for $10$ epochs, using the transferred shared model. 
For CIFAR-10, we randomly selected 10 clients as new clients and trained the shared model using the remaining 90 clients, including 20 Byzantine clients. Similarly, for FEMNIST, we selected 10 new clients and trained the shared model with the remaining 140 clients, of which 50 were Byzantine.
We evaluated the average accuracy for four approaches. 
To highlight the benefits of MTRL for transfer learning, we considered a Naive setting, where each client trains their model independently, resulting in poor accuracy. Figures~\ref{fig:metacfsr}, \ref{fig:metaemsr}, \ref{fig:metaml} and~\ref{fig:metasr} illustrate the results: FedRep, without Byzantine attacks, demonstrates effective transferability of the shared model to new clients. In contrast, Byzantine FedRep, with malicious clients with two type of attacks, exhibits significantly lower performance, highlighting the vulnerability of FL systems. Finally, BR-MTRL, our proposed approach, improves learning accuracy by employing GM and {\cblue Krum} aggregation, validating its effectiveness u under Byzantine attacks.
%
\section{Conclusion}
In this paper, we proposed a Byzantine-resilient federated MTRL framework to learn robust shared representations across {\em related yet different} tasks. Our approach employs an alternating gradient descent strategy, where the shared model is updated while keeping the local head fixed, and vice versa. To ensure Byzantine resilience, we incorporated geometric median and Krum-based aggregation at the server, mitigating the impact of malicious clients sending corrupted updates. We implemented our federated learning approach on AWS, simulating a real-world-like federated setting. Numerical experiments on CIFAR-10 and FEMNIST validate the effectiveness of our approach and its transferability to new clients with limited data.
%
\bibliography{Byz}
\bibliographystyle{IEEEtran}
\end{document}




\end{document}